\documentclass[letterpaper, 10 pt, journal, twoside]{ieeetran}  

\IEEEoverridecommandlockouts                              

\usepackage[utf8x]{inputenc}
\usepackage{amsmath}
\usepackage{amssymb}
\usepackage{wrapfig}
\usepackage{graphicx}
\usepackage{algorithm}
\usepackage{algpseudocode}
\usepackage{eqparbox}
\usepackage{mdwmath}
\usepackage{mdwtab}
\usepackage{array}
\usepackage{authblk}
\usepackage{bm}

\usepackage{hyperref}
\usepackage{units}
\usepackage{multirow}

\usepackage{subfigure}

\usepackage{setspace}

\usepackage[left=2cm,right=2cm,top=2cm,bottom=2cm]{geometry}

\usepackage[colorinlistoftodos, textwidth=3.7cm]{todonotes}

\usepackage[english]{babel}

\usepackage{pslatex}

\graphicspath{
	{figs/}
	{matlab/}
	{anim/}
}

\renewcommand{\vec}[1]{\mathbf{#1}}

\title{\LARGE 
ObserveNet Control: A Vision-Dynamics Learning Approach to Predictive Control in Autonomous Vehicles
}

\author{
Cosmin~Ginerica$^{1}$,
Mihai~Zaha$^{1}$,
Florin~Gogianu$^{2}$,
Lucian~Busoniu$^{3}$,
Bogdan~Trasnea$^{1}$,
and~Sorin~Grigorescu$^{1}$
\thanks{
Manuscript received: February 24, 2021; Revised: May 28, 2021; Accepted: July 6, 2021.}
\thanks{
This paper was recommended for publication by Editor Jens Kober upon evaluation of the Associate Editor and Reviewers' comments.}
\thanks{
$^{1}$ Cosmin Ginerica, Mihai Zaha, Bogdan Trasnea and Sorin Grigorescu are with \href{http://www.rovislab.com/}{RovisLab (Robotics, Vision and Control Laboratory, Romania)}}
\thanks{
$^{2}$ Florin Gogianu is with \href{https://www.bitdefender.ro/}{BitDefender}, Romania}
\thanks{
$^{3}$ Lucian Busoniu is with Technical University of Cluj-Napoca, Romania
}
\thanks{
Digital Object Identifier (DOI): see top of this page.}
}

\begin{document}

\maketitle
\begin{abstract}

A key component in autonomous driving is the ability of the self-driving car to understand, track and predict the dynamics of the surrounding environment. Although there is significant work in the area of object detection, tracking and observations prediction, there is no prior work demonstrating that raw observations prediction can be used for motion planning and control.

In this paper, we propose \textit{ObserveNet Control}, which is a vision-dynamics approach to the predictive control problem of autonomous vehicles. Our method is composed of a \textit{i}) deep neural network able to confidently predict future sensory data on a time horizon of up to $10s$ and \textit{ii}) a temporal planner designed to compute a safe vehicle state trajectory based on the predicted sensory data. Given the vehicle's historical state and sensing data in the form of Lidar point clouds, the method aims to learn the dynamics of the observed driving environment in a self-supervised manner, without the need to manually specify training labels. The experiments are performed both in simulation and real-life, using CARLA and RovisLab's AMTU mobile platform as a 1:4 scaled model of a car. We evaluate the capabilities of ObserveNet Control in aggressive driving contexts, such as overtaking maneuvers or side cut-off situations, while comparing the results with a baseline Dynamic Window Approach (DWA) and two state-of-the-art imitation learning systems, that is, Learning by Cheating (LBC) and World on Rails (WOR).

\end{abstract}


\section{Introduction}
\label{sec:introduction}

\IEEEPARstart{W}{ith} the rise of Artificial Intelligence and Deep Neural Networks (DNN), research in the area of autonomous vehicles has advanced significantly in the last decade, being driven both by academia and industry alike. An autonomous car is an intelligent agent which observes its environment, makes decisions and performs actions based on these decisions. The state-of-the-art approaches for self-driving car control are based on the traditional perception-planning-act pipeline~\cite{pendleton2017perception}, where first the environment is reconstructed from sensory data. Secondly, a safe vehicle state trajectory path is computed by a path planner. Finally, the path is translated to motion commands which are executed by a controller.

A key difficulty is that the control strategy is affected by unpredictable external factors that can arise in the driving environment. It has been observed that the control quality increases when the motion of the objects around the car is tracked and predicted. However, such an approach is strictly dependent on the accuracy of the object detection system. If the detection system fails, than the motions of the objects cannot be predicted anymore, hence degrading the performance of the controller.

\begin{figure}
\centering
	\begin{center}
		\includegraphics[scale=0.95]{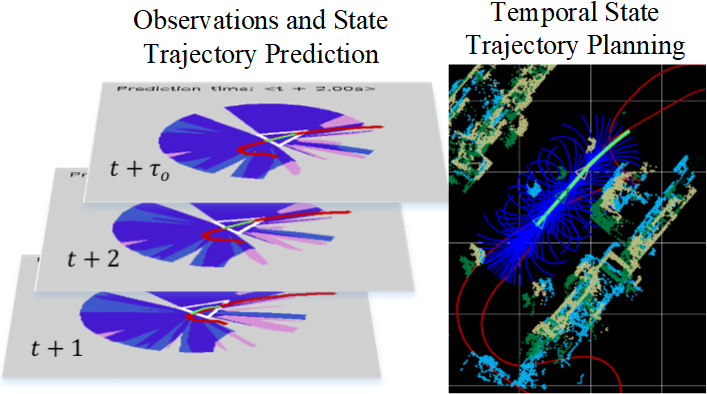}
		\caption{\textbf{ObserveNet Control.} A deep neural network is trained in a self-supervised fashion to predict raw sensory data over a finite prediction horizon $[t+1, t+\tau_o]$. The predictions, as well as historic states and observations, are iteratively used by a temporal state trajectory planner to calculate and execute a safe vehicle motion.}
        \label{fig:concept}
	\end{center}
    \vspace{-2em}
\end{figure}

In this paper, we propose the \textit{ObserveNet Control} method depicted in Fig.~\ref{fig:concept}, which is an approach to derive control signals from predicted raw sensory data. The algorithm is composed of a DNN which receives as input historical system states and measurements from a Lidar range sensing device. The network is composed of convolutional and recurrent layers trained in a self-supervised fashion, without the need of manually labeling training data. The output of the DNN represents the most likely future state trajectory of the vehicle, where each state in the state trajectory is augmented with predicted sensory data.

Our work was mainly inspired by~\cite{world_discovery_models}, which served as a baseline for developing the architecture of the DNN. Apart from the application area (autonomous driving vs. navigating a simulated maze), we highlight the several key differences in our work, as detailed in Section~\ref{sec:setup}. For performance evaluation, we have tested our algorithm on two distinct platforms, namely within the CARLA simulator, as well as on our RovisLab Autonomous Mobile Testing Platform (AMTU) from Fig~\ref{fig:model_car}. 

The key contributions of the paper are:

\begin{itemize}
	\item the deep neural network architecture of ObserveNet, used to predict future state trajectories and raw sensory measurements along a finite prediction horizon;
    \item the self-supervised training of ObserveNet with raw range sensing information acquired from a Lidar device;
    \item a temporal planner used to calculate a safe vehicle path based on the output of ObserveNet.
\end{itemize}

The rest of the paper is organized as follows. Section~\ref{sec:related_work} covers the related work. The methodology of ObserveNet Control is given in Section~\ref{sec:method}. The experimental validation is given in Section~\ref{sec:experiments}. Finally, conclusions are stated in Section~\ref{sec:conclusions}.

\section{Related Work}
\label{sec:related_work}

Similar to the perception-planning-act \cite{pendleton2017perception} paradigm, our approach divides the algorithm into raw observations prediction and state trajectory planning. This is opposed to End-To-End learning~\cite{bojarski2016end}, where sensory data is directly mapped to control signals via a deep neural network.


As a result of progress in AI, several model-free approaches have emerged in the field of learning-based control for self-driving vehicles. One such paradigm is End-To-End Learning, where input data is mapped directly to control signals. Another approach is Deep Reinforcement Learning (DRL), where an agent is controlled via action-reward systems~\cite{Jaritz_ICRA2018}. In terms of training data, self-supervised learning is a technique heavily used in modern learning-based architectures, consisting of algorithms that learn independently, without the need of human intervention in the form of manual data labelling.

From a data-driven control perspective, our work builds on goal-conditioned imitation learning~\cite{C, D, F, chen2021wor}. As shown in~\cite{D}, this class of methods combines learning from an expert driver with the advantages of goal-directed planning based on dynamic models and reward functions. However, in addition learning the optimal driving path, we also employ the prediction of future raw observations within the control task.

Additional to accurate object detection, the performance of autonomous vehicles is directly influenced by their ability to predict future positions of dynamic agents, such as other cars and pedestrians~\cite{G}. Instead of focusing on label-intensive object detection and tracking, we proposes to learn to predict future sensory data via self-supervised learning and plan future trajectories based on the predicted sensory data. Raw observations predictions include PredNet~\cite{Lotter2016DeepPC}, which was trained to predict future frames in a video sequence, while the work in~\cite{H, I, K, L} is focused on forecasting an intermediate occupancy map representation of the surrounding traffic environment. An inverted pose forecasting pipeline is proposed in~\cite{J}, where the first phase of the algorithm predicts future 3D point clouds, followed by object detection and pose estimation, instead of first detecting objects and then predicting their future motion. Apart from sensory data, LaneGCN~\cite{LaneGCN} and VectorNet~\cite{VectorNet} are used to fuse object detection and tracking with HD map data in order to better forecast the motion of dynamic objects within complex road networks. However, using our ObserveNet sensory data predictor, we also propose to derive control signals for the vehicle, thus showing that automatic control based on predicted observations is viable.


The authors of \cite{world_discovery_models} use a temporal neural network to construct an agent's belief. The belief is used in a RL setup to build a world model. The agent's observations are modeled as one-hot encoded obstacles in an occupancy grids. The goal is to learn an information seeking behavior for an agent, endowing it with a strong drive towards discovering its environment. In our work, we use a different observation modelling mechanism for Lidar data. While the agent in~\cite{world_discovery_models} moves based on four discrete control actions (left, right, up, down), in our work the actions are represented by vehicle control signals. Another difference is that the authors in \cite{world_discovery_models} use a control method based on Reinforcement Learning, as opposed to our ObserveNet Control pipeline.

A key contribution of our work is therefore to extend the neural network architecture from \cite{world_discovery_models}, adapt it to our sensor setup, and illustrate that this modified architecture can provide results in a closed-loop system, tested under dynamic traffic conditions and with a real-world robotic system.

\section{Methodology}
\label{sec:method}

Throughout the paper, we use the following notation. The value of a variable is defined either for a single discrete time step $t$, written as superscript $<t>$, or as a discrete sequence defined in the $<t, t+k>$ time interval, where $k$ denotes the length of the sequence. For example, the value of a state variable $\vec{z}$ is defined either at discrete time $t$ as $\vec{z}^{<t>}$, either within a sequence interval $\vec{z}^{<t, t+k>}$. The difference between a measured and a predicted quantity is made using the hat notation(e.g $\vec{z}$ and $\hat{\vec{z}}$ are measured and predicted values, respectively). Vector and matrices are indicated by bold symbols.

\subsection{Problem definition}
\label{problem_definition}

The autonomous driving problem can be stated as follows. Given a global reference trajectory $\vec{z}^{<t-\infty, t+\infty>}_{ref}$, we want to calculate and execute a safe vehicle state trajectory $\vec{z}^{<t+1, t+\tau_o>}$, given past and current sensory observations $\vec{I}^{<t-\tau_i, t>}$, as well as predicted raw observations $\vec{\hat{I}}^{<t+1, t+\tau_o>}$. $t$ is the discrete time, while $\tau_i$ and $\tau_o$ are the length of the sequence of past observations and the control horizon, respectively.

A state trajectory $\vec{z}^{<t-i, t+i>}$ is defined as a sequence of vehicle states:

\begin{equation}
    \label{eq:trajectory}
    \vec{z}_{ref} = [x, y, v, \phi]
\end{equation}

\noindent where $x$ and $y$ are the Cartesian coordinates of the ego-vehicle, $v$ is the velocity and $\phi$ the orientation.

\subsection{ObserveNet Predictive Control}

\begin{figure}
\centering
	\begin{center}
		\includegraphics[scale=0.9]{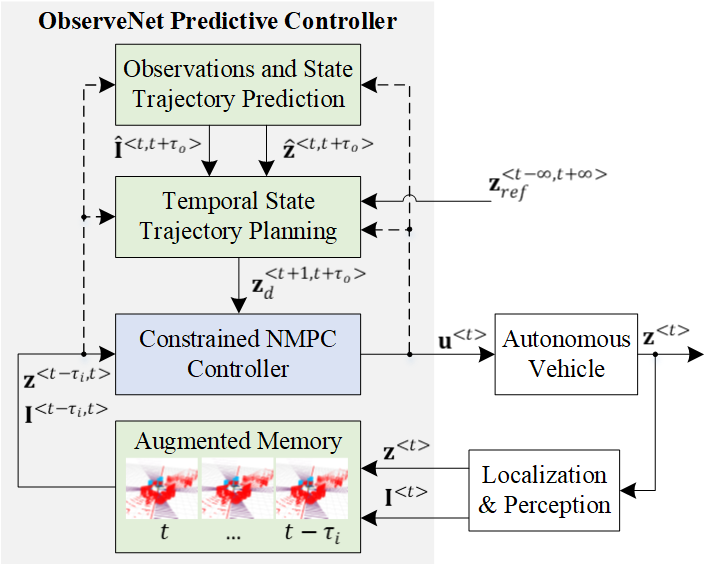}
        \vspace{-1em}
		\caption{\textbf{Block diagram of the ObserveNet Predictive Control algorithm.} Based on historical Localization \& Perception data stored in the Augmented Memory component, the method predicts raw sensory Observations and State Trajectories which are used by a Temporal Planner to calculate a safe desired vehicle trajectory executed via Constrained Non-linear Model Predictive control. The dotted lines illustrate the flow of data used during self-supervised training.}
        \label{fig:setup}
	\end{center}
    \vspace{-2em}
\end{figure}

The objective of the control loop is to compute a safe state trajectory $\vec{z}^{<t+1, t+\tau_o>}_{d}$ which is to be executed by a motion controller. Fig~\ref{fig:setup} illustrates our proposed ObserveNet predictive control approach. Our main contributions are the Observations and Trajectory Prediction module, along with the Temporal State Trajectory Planning algorithm which consumes past and predicted observations and states, while computing the input to a Constrained Non-linear Model Predictive (NMPC) controller. Within its optimization loop, NMPC calculates optimal the control signals $\vec{u}^{<t>} = (v^{<t>}_{cmd}, \delta^{<t>}_{cmd})$, where $v^{<t>}_{cmd}$ and $\delta^{<t>}_{cmd}$ are the longitudinal velocity and steering angle, respectively.

The Localization \& Perception component from Fig.~\ref{fig:setup} constructs a top-view 2D occupancy grid out of 3D Lidar point clouds, while computing the state of the vehicle using inertial measurements from an IMU, geolocation data from a GPS receiver and wheels odometry. The Ackermann vehicle model is used both by the state estimator and the controller:

\begin{equation}
	\hat{\vec{z}}^{<t+1>} = \vec{z}^{<t>} + v^{<t>} \cdot
		\begin{bmatrix}
			\cos (\phi^{<t>}) \\
			\sin (\phi^{<t>}) \\
            1 \\
			\frac{1}{L} \tan (\delta^{<t>})
		\end{bmatrix}
		\Delta t
	\label{eq:nominal_process_model}
\end{equation}

\noindent where $\delta$ is the steering angle, $L$ is the length of the vehicle and $\Delta t$ is the sampling time.

The predictions are integrated by taking into consideration the vehicle's dynamics from Eq.~\ref{eq:nominal_process_model}, so that obstacle information is projected onto the occupancy grid according to the ego-vehicle's state for each predicted set of observations. In the following, we detail ObserveNet's deep neural network architecture and the methodology within the temporal state trajectory planner.

\subsection{ObserveNet Architecture and Training}

The goal of the ObserveNet neural network from Fig.~\ref{fig:setup} is to predict future observations $\vec{\hat{I}}^{<t+1, t+\tau_o>}$ based on past sensory measurements $\vec{I}^{<t-\tau_i, t>}$, the global reference route and the past vehicle's state trajectory $\vec{z}^{<t-\tau_i, t>}$. This past and predicted data is used to calculate the reference of the constrained NMPC controller which will execute the motion of the car.

\begin{figure*}
\centering
	\begin{center}
		\includegraphics[scale=0.9]{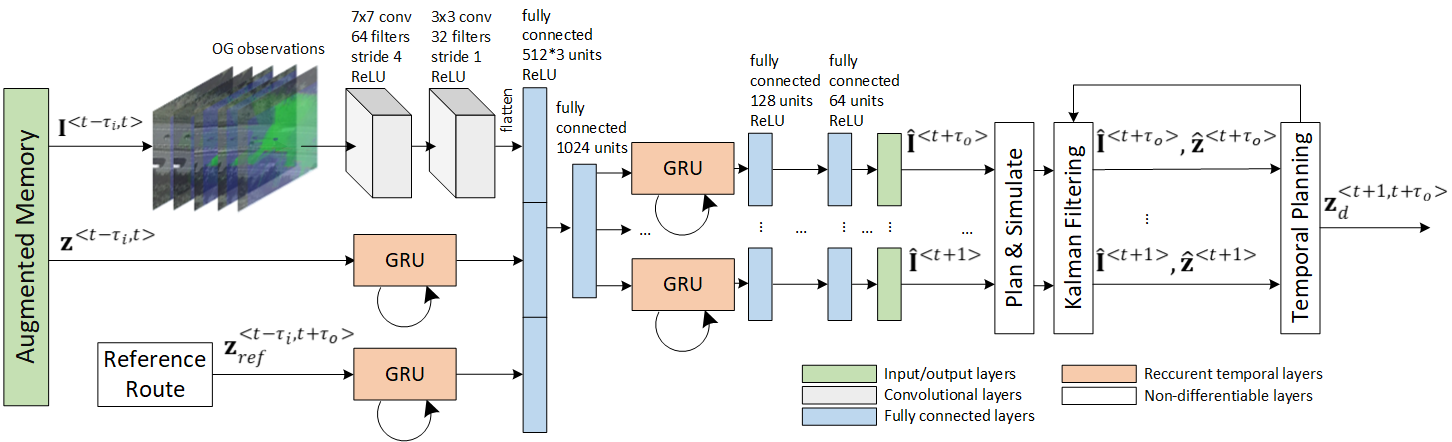}
        \vspace{-1em}
		\caption{\textbf{ObserveNet deep neural network architecture.} The DNN receives input data in the form of historical observations, vehicle states and reference route, with the goal of computing a safe desired state trajectory $\vec{z}^{<t+1, t+\tau_o>}_d$ using predicted observations and temporal path planning. ObserveNet computes a dynamic output, having a size dependent on the time horizon for which the future observations are predicted.}
        \label{fig:network_architecture}
	\end{center}
    \vspace{-1em}
\end{figure*}

In order to increase the computational speed, the input Lidar point clouds are firstly converted into an OcTree representation. An octree is a data structure used for representing 3D spatial features by subdividing them recursively into eight child units. The obtained 3D grid map can be calculated at a finer or more coarse resolution, while preserving a low-memory footprint, making the representation suitable for embedded computing. Within the conversion loop, we use several downsampling mechanisms, as follow.

First, we introduced a constraint for points which lay farther than a distance threshold:

\begin{equation}
    \label{eq:point_cloud_threshold}
    p_r = p[i], \iff d(s, p[i]) < T_d, \forall i \in [1, N]
\end{equation}

\noindent where $p[i]$ and $p_r$ are input and output point clouds, respectively. $d(s, p[i])$ is the Euclidean distance between the ego-vehicle's origin and the current point, $T_d$ is the distance threshold and $N$ is the original point cloud size.

Secondly, points which are below a certain height or higher than our ego-vehicle's height are discarded:

\begin{equation}
    \label{eq:point_cloud_high_points}
    p_r = p[i], \iff V_L \leq p[i]_Z \leq V_H, \forall i \in [1, N]
\end{equation}

\noindent where $p[i]_Z$ is height of point, $V_L = 0.25m$ is the minimum height, $V_H$ is the ego-vehicle height and $N$ is the point cloud size. The goal of $V_L$ is to filter our the pointclouds representing the road.

Finally, points which are outside of an $160^\circ$deg field of view from the front and rear of the vehicle are filtered out:

\begin{equation}
    \label{eq:point_cloud_fov}
    p_r = p[i] \iff |\angle(p[i], V_O)| \leq T_v \forall i \in [1, N]
\end{equation}

\noindent where $V_O$ is the vehicle origin, $T_v$ is the field of view threshold and $N$ is the point cloud size. The problem of slopes, when the road would be visualized as obstacles in the pre-processed pointcloud, can be easily counteracted by updating the extrinsic parameters of the Lidar, which in turn relate its orientation with respect to the vehicle.

The obtained OcTree is stored as historical data in the Augmented Memory component from Fig.~\ref{fig:setup}. In order to further increase the computational speed on ObServeNet, we project the OcTree onto a top-view 2D occupancy grid which is feed to the deep network. The network outputs observation predictions for $\tau_o$ future frames.

ObserveNet's architecture is shown in Fig.~\ref{fig:network_architecture}. The basic building blocks are convolutional layers for spatial data encoding and Gated Recurrent Units (GRU) for extracting temporal dependencies. The input sequences are normalized and fed through the network's convolutional layers. The proposed DNN has a dynamic output, dependent on the length of the prediction horizon $\tau_o$. Namely, each predicted observation $\hat{\vec{I}}^{<t+k>}$ is given by a separate network output branch $k$, where $k \leq \tau_o$.

\begin{figure}
\centering
	\begin{center}
		\includegraphics[scale=0.9]{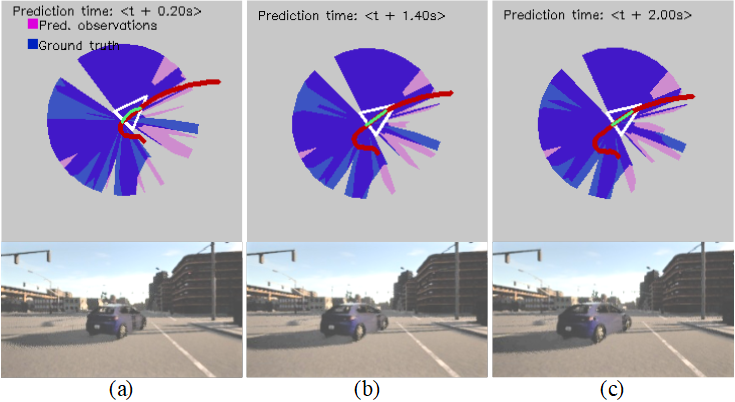}
        \vspace{-0.5em}
		\caption{\textbf{Predicted and ground truth observations at three different future timesteps.} (top) Ground truth observations superimposed on ObserveNet predictions, along with the reference path marked in red and the planned state trajectory illustrated in green. For visualization purposes, we have considered only observations within a $50m$ range. (bottom) Images from the front of the vehicle at the considered timestamps.}
        \label{fig:observenet_predictions}
	\end{center}
    \vspace{-2em}
\end{figure}

As illustrated in Fig.~\ref{fig:network_architecture}, apart from the differentiable layers in the network, that is, its convolutional, recurrent and dense units, the architecture also embodies non-differentiable layers used to map each predicted observation to the most probable future state of the vehicle. The Plan \& Simulate layer recursively uses our temporal planner to simulate future states, while replanning the vehicle desired trajectory $\vec{z}^{<t+1, t+\tau_o>}_d$. The predicted states are smoothed using an extended Kalman filter. A couple of observations predictions an their ground truth are shown in Fig.~\ref{fig:observenet_predictions}, as well as in Fig.~\ref{fig:temporal_planner} in the context of the temporal planner.

ObserveNet has been trained for $10.000$ epochs in a self-supervised fashion using the Adam optimizer, with learning rate of 0.0003, while minimizing the mean squared error loss function between measured and predicted observations.

\subsection{Temporal State Trajectory Planning and Control}

The temporal path planner is based on the Dynamic Window Approach (DWA) method, modified to iteratively compute optimal vehicle paths based on current, as well as predicted observations. The Dynamic Window Approach (DWA) is an online collision avoidance strategy for mobile robots, which uses robot dynamics and constraints imposed on the robot's velocities and accelerations to calculate a collision free path in the 2D plane. We have implemented DWA based on the Robot Operating System (ROS) DWA local planner. DWA takes as input the distances from the ego-vehicle to the obstacles present in the scene, calculated as perception rays. In our case, we tune DWA to take into consideration both past, as well as future predicted information.

\begin{figure}
\centering
	\begin{center}
		\includegraphics[scale=0.7]{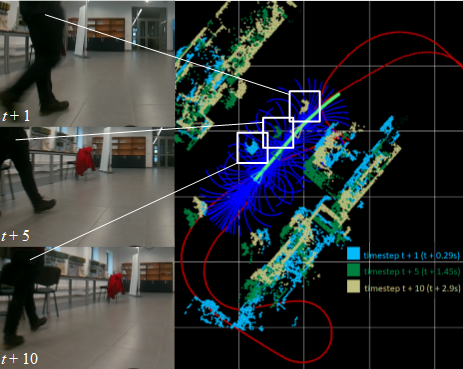}
        \vspace{-1em}
		\caption{\textbf{Temporal state trajectory planning using RovisLab AMTU.} While tracking the red reference path, ObserveNet predicts the future observations marked with blue, green and yellow, respectively, as well as the most probable state trajectory of the vehicle. The blue lines represent candidate trajectories at each future timestamp, while the green line represents the output of the temporal planner. The white rectangles show predicted observations of the person depicted on the left, considered in this case as a moving obstacle.}
		\label{fig:temporal_planner}
	\end{center}
    \vspace{-2em}
\end{figure}

A desired vehicle state is iteratively predicted at each time step in the future, based on the temporal vehicle dynamics and observations. The input data is both historical ($<t-\tau_i, t>$), as well as predicted ($<t, t-\tau_o>$):

\begin{equation}
    \label{eq:trajectory_generation}
    \vec{z}^{<t+i>} = f(\hat{\vec{I}}^{<t-\tau_i, t+i>}, \vec{z}^{<t+i-1>})
\end{equation}

\noindent where $i$ is the prediction timestep and $f({\cdot})$ if the temporal path planning function. 
$f({\cdot})$ is used to interpolate the outputs of the DWA planner along the time horizon $[t-\tau_i, t+\tau_o]$ and can be decomposed as:

\begin{equation}
    \label{eq:temporal_path_planner_function}
    f({\cdot}) = \sum_{t=-\tau_i}^{0} DWA(\vec{I}^{<t>}, \vec{z}^{<t>}) + \sum_{t=1}^{\tau_o} DWA(\hat{\vec{I}}^{<t>}, \hat{\vec{z}}^{<t>})
\end{equation}

\noindent where the first term represents the interpolated trajectories of the planner given past observations, while the second term computes the interpolation for the planned trajectories based on predicted observations. The states are iteratively integrated into the desired state trajectory $\vec{z}^{<t+1, t+\tau_o>}_d$ feed as input to the NMPC.

A snapshot of our temporal path planner can be seen in Fig.~\ref{fig:temporal_planner}. The planner generates candidate trajectories at each prediction step (shown with blue lines in Fig.~\ref{fig:temporal_planner}). The optimal trajectory is selected based on the vehicle dynamics from Eq.~\ref{eq:nominal_process_model} and predicted observations. The final trajectory computed using Eq.~\ref{eq:temporal_path_planner_function}, marked with green in Fig.~\ref{fig:temporal_planner}.

\section{Experiments}
\label{sec:experiments}

The performance of ObserveNet Control was benchmarked against the classical Dynamic Window Approach (DWA) baseline method, as well as against two state-of-the-art methods based on imitation learning, that is, Learning by Cheating (LBC)~\cite{B} and World on Rails (WOR)~\cite{chen2021wor}. Since ObserveNet is a derivation of imitation learning, we have chosen to compare it to LBC and WOR, which are methods that do not require object-bound boxes, tracks and HD maps, like~\cite{LaneGCN} or~\cite{VectorNet}. DWA and ObserveNet use the same Constrained NMPC controller for executing the motion of the vehicle.

\begin{figure}
\centering
	\begin{center}
		\includegraphics[scale=1.0]{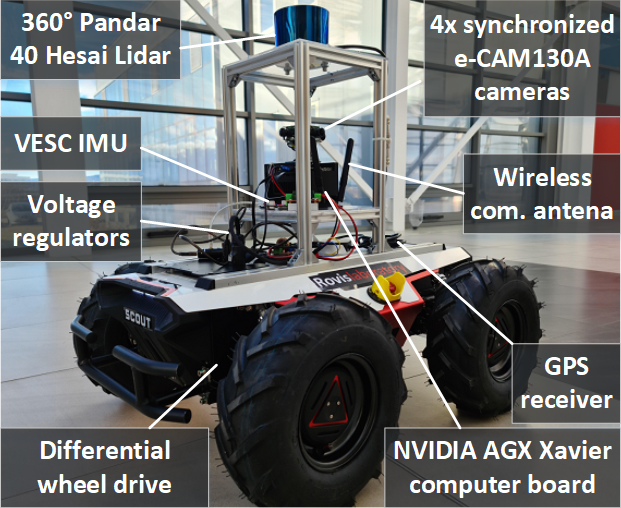}
		\caption{\textbf{RovisLab AMTU (Autonomous Mobile Test Unit).} The platform is used in the context of a 1:4 scaled car model.}
        \label{fig:model_car}
	\end{center}
    \vspace{-2em}
\end{figure}

The evaluation of the four competing algorithms was performed on aggressive driving scenarios simulated in the city environments Town 1 to 6 in CARLA, as well as on CARLA Leaderboard~\cite{A}. Real-world experiments have been performed using DWA and ObserveNet on indoor and outdoor navigation using the RovisLab Autonomous Mobile Test Platform (AMTU) from Fig.~\ref{fig:model_car}.

Global reference trajectories were recorded while controlling both simulated and real vehicles manually and used afterwards as reference trajectories for the autonomous vehicle controller to track.

As performance measures, we have recorded the number of collisions $N_c$ between the ego-vehicle and other traffic participants, the cross-track error $e_{ct}$, the vehicle's orientation $e_{\phi}$, the rate of change of the steering and acceleration commands $e_{\delta}$ and $e_{a}$, respectively, as well as the ability of the vehicle to successfully complete the pre-recorded track, quantified as $C_{100\%}$.

The cross-track error represents the difference between the reference trajectory and the vehicle's position along the $y$ coordinate:

\begin{equation}
    \label{eq:cte}
    e_{ct} = g(x) - y
\end{equation}

\noindent where $g(x)$ represents the polynomial approximation of the trajectory, evaluated at $x$.

The vehicle orientation error is defined as:

\begin{equation}
    \label{eq:epsi}
    e_{\phi} = \phi - \phi_{d} + \frac{v}{L}\delta 
\end{equation}

\noindent where $\phi$ is the vehicle orientation, $\phi_d$ is the desired vehicle orientation, $v$ is the ego-vehicle velocity, \textit{L} is the vehicle wheelbase and $\delta$ is the steering angle.

The rate of change of the steering and acceleration commands is computed as:

\begin{equation}
    \begin{cases}
    e_{\delta} &= \delta^{<t+1>} - \delta^{<t>} \\
    e_{a} &= a^{<t+1>} - a^{<t>}
    \end{cases}
\end{equation}

\begin{figure*}
\centering
	\begin{center}
		\includegraphics[scale=0.82]{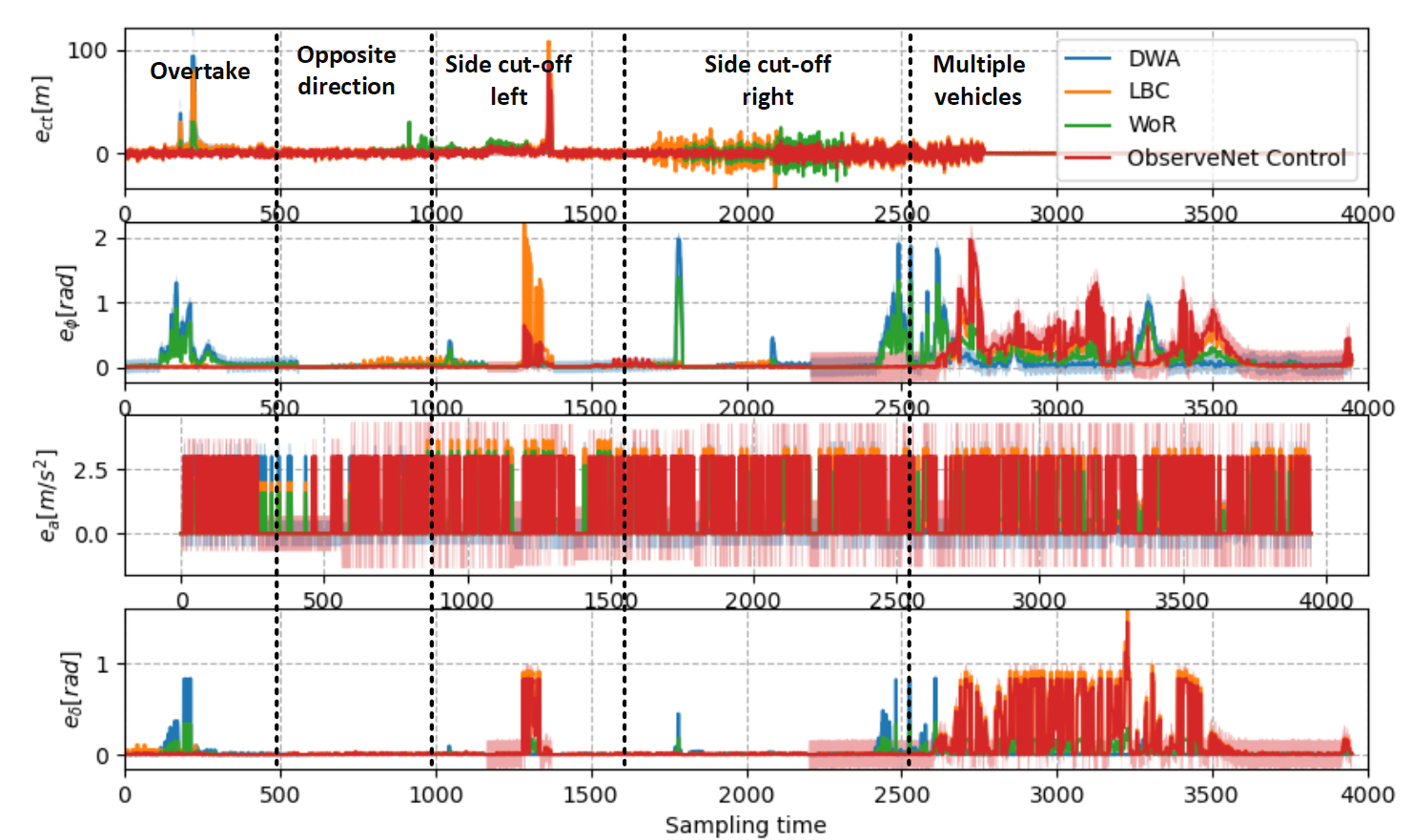}
        \vspace{-1em}
		\caption{\textbf{Mean (solid line) and standard deviation (shaded region) of the performance metrics recorded during testing in the five considered scenarios.} Per sampling time cross-track error, orientation error, consecutive acceleration difference and consecutive steering angle difference analysis. From left to right: overtake, opposite direction, side cut-off left, side cut-off right and multiple vehicle scenarios statistics.}
		\label{fig:statistics_combined}
	\end{center}
    \vspace{-2em}
\end{figure*}

The workflow of the experiments is as follows:

\begin{itemize}
    \item collect training data from driving recordings;
    \item format training data as historical sequences of length $N$ and prediction sequences of length $\tau_o$;
    \item train the ObserveNet deep network from Fig.~\ref{fig:network_architecture};
    \item evaluate on simulated and real-world driving scenarios.
\end{itemize}

To train the network, we use data collected from CARLA and RovisLab AMTU, with the ego-vehicle being manually driven in the test environment, while encountering various traffic participants. In order to to mitigate overfitting, all vehicles were driven in a different manner each time. During training, we have used an 80/10/10 train - validation - test data split.

\subsection{Simulation Experiments}
\label{sec:setup}

CARLA \cite{Dosovitskiy17} was used for testing ObserveNet Control in simulation. We have simulated a $360^\circ$ Lidar sensor, spanning a maximum range of $50$m. Additionally, camera images, inertial data and odometry information was acquired.

Due to the high flexibility of the CARLA simulator, we were able to simulate real-world situations like overtaking, opposite direction lane swerving, side cut-off, etc. During the evaluation step, our algorithm was used to control the ego-vehicle, while the other traffic participants were controlled using a simple feedback loop directly in CARLA.

Towns 1-6 from CARLA are used for training and testing, amounting to $90$K pairs of Lidar and vehicle state data. We have investigated city driving with two or more simulated vehicles driving in the environment. We tackle aggressive driving, mainly overtaking and cut-off situations, where a classic controller would have difficulties due to the scene's high dynamics. The ego-vehicle navigated a pre-recorded trajectory, while several challenging conditions were added. Other traffic participants were simulated as follows: \textit{i}) travelling at a dangerously low speed, leading to the need of aggressive overtaking maneuvers, \textit{ii}) vehicles incoming from the opposite direction at an unsafe distance and \textit{iii}) vehicles cutting off the ego-vehicle from the left and right sides.

The goal of our experiments was to check if we can eliminate collisions, as well as to smooth the state trajectory, thus generating a safer more pleasant ride, with less jerking. We have implemented five challenging scenarios, with quantitative results shown in Fig.~\ref{fig:statistics_combined}.

The \textit{overtake scenario} represents a challenging situation, particularly at higher speed. The other vehicle accelerates at a slower pace than the ego-vehicle, towards a lower top speed, generating an overtake situation. State data is recorded for generating statistics.

In the \textit{opposite direction} experiment, the ego-vehicle travels straight, while the other vehicle is coming from the opposite direction, spawning a near-miss situation. Using the network's predictions, the ego-vehicle should sense the other vehicle sooner than the setup without predictions, generating a safer trajectory.

The \textit{side cut-off} scenario consists of the ego-vehicle travelling straight and approaching an intersection. When it comes close to the intersection, another vehicle cuts it off, generating a collision. Using the predicted data, the ego-vehicle needs to be able to avoid the collision. This scenario was tested in two symmetrical situations: left side cut-off and right side cut off.

The \textit{rear pass-by} scenario simulates a vehicle coming from behind the ego-vehicle on the same side, at a greater velocity, passing it at an unsafe distance. Using the predictions, the other vehicle's position is predicted, thus allowing the ego-vehicle's path planner to plan accordingly, widening the gap between the vehicles.

The \textit{multi-vehicle} scenario has the most complexity, where the ego-vehicle is traveling a larger map section, encountering vehicles, overtaking and avoiding collisions. The scenario is designed to simulate real world complex scenes, where the ego-vehicle has limited maneuver space.

For all five experimental scenarios, the corresponding quality measures are summarized in Table \ref{tab:results}. The experiments show that ObserveNet achieves better results than the regular DWA planner, providing less jerky movements. This is due to the fact that the obstacles are available before being actually observed.

The ego-vehicle successfully avoided obstacles in all five experimental scenarios. The cross-track error remained relatively low, although in some circumstances it was greater than the DWA baseline for all three imitation learning methods (e.g. when running the \textit{side cut-off} or \textit{multi-vehicle} scenarios). This behavior happens when the ego-vehicle is overtaking, thus deviating from the reference trajectory. We believe that the lower performance of ObserveNet, LBC and WOR against DWA in the \textit{side cut-off} and \textit{multi-vehicle} scenarios is due to the randomness of these scenarios, as well as due to the high dynamics of the involved participants. An additional amount of training data should increase performance.

The overtaking maneuvers also cause spikes in orientation error levels. When running the overtake scenario solely with DWA, the ego-vehicle collided violently with the other traffic participant, thus causing large cross-track and orientation errors. Consecutive acceleration and steering angle differences were in some instances better without predictions, however the error values were very close.

\begin{table}[]
    \centering
    \resizebox{\columnwidth}{!}
    {
    \begin{tabular}{ccccccc}
        \hline
        \textbf{Scenario} & \textbf{Algorithm} & \textbf{$N_C$} & \textbf{$\bar{e_{ct}}$} & \textbf{$\bar{e_{\phi}}$} & $\bar{e_{a}} / \bar{e_{\delta}}$ & \textbf{$C_{100\%}$}\\
        \hline
        \multirow{4}{*}{Overtake}           & DWA  & 1 & \textbf{0.06} & 0.02 & 0.001 / 0.001 & 100 \\
                                            & LBC  & \textbf{0} & 0.076 & 0.027 & 0.0069 / 0.0023 & 100 \\
                                            & WOR  & \textbf{0} & 0.065 & 0.019 & 0.0061 / 0.0012 & 100 \\
                                            & Ours & \textbf{0} & 0.6 & \textbf{0.00001} & \textbf{0.0006} / \textbf{0.0006} & 100 \\
        \hline
        \multirow{4}{*}{Opposite}           & DWA  & 1 & 0.72 & 0.003 & 0.006 / 0.001 & 100 \\
                                            & LBC  & \textbf{0} & 0.2 & \textbf{0.0023} & 0.042 / 0.005 & 100 \\
                                            & WOR  & \textbf{0} & 0.17 & 0.04 & 0.031 / 0.01 & 100 \\
                                            & Ours & \textbf{0} & \textbf{0.13} & 0.004 & \textbf{0.0008} / \textbf{0.0008} & 100 \\
        \hline
        \multirow{4}{*}{Side cut-off}       & DWA  & 1 & \textbf{0.002} & 0.06 & \textbf{0.002} / 0.002 & 100 \\
                                            & LBC  & 1 & 0.10 & 0.23 & 0.064 / 0.020 & 100 \\
                                            & WOR  & \textbf{0} & 0.05 & 0.04 & 0.0096 / 0.010 & 100 \\
                                            & Ours & \textbf{0} & 0.003 & \textbf{0.005} & 0.004 / \textbf{0.001} & 100 \\
        \hline
        \multirow{4}{*}{Rear pass-by}       & DWA  & N/A & 1.57 & 0.007 & 0.01 / \textbf{0.003} & 100 \\
                                            & LBC  & N/A & 0.45 & 0.02 & 0.063 / 0.020 &  100 \\
                                            & WOR  & N/A & 0.62 & \textbf{0.005} & 0.0067 / 0.010 & 100 \\
                                            & Ours & N/A & \textbf{0.39} & 0.02 & \textbf{0.004} / 0.004 & 100 \\
        \hline
        \multirow{4}{*}{Multi-vehicle}      & DWA  & 4 & \textbf{0.005} & 0.04 & 0.0003 / \textbf{0.0003} & 24.56 \\
                                            & LBC  & 1 & 0.006 & 0.34 & \textbf{0.0002} / 0.0004 & 100 \\
                                            & WOR  & 1 & 0.006 & 0.13 & 0.00047 / 0.0011 & 100 \\
                                            & Ours & \textbf{0} & 0.01 & \textbf{0.01} & 0.0005 / 0.0005 & 100  
    \end{tabular}
    }
    \caption{Quantitative results on simulated scenarios in CARLA.}
    \label{tab:results}
    \vspace{-3em}
\end{table}

\subsection{CARLA Leaderboard Experiments}

On top of our own CARLA simulation experiments, we have also assessed ObserveNet within Carla Autonomous Driving Leaderboard \cite{A}, which is an open platform designed to ease testing of autonomous driving systems. The platform provides several urban environments, as well as varying weather and lighting conditions and can be configured to simulate a variety of challenging traffic scenarios, such as vehicle control loss, obstacle avoidance, lane changing, crossing traffic running a red light and intersection handling. Three metrics are used for performance evaluation: total driving score, route completion and infraction penalty. In addition to these metrics, the framework provides collision information, off-road driving data, route deviations and running red lights and stop signs.

\begin{figure}
\centering
	\begin{center}
		\includegraphics[width=\columnwidth]{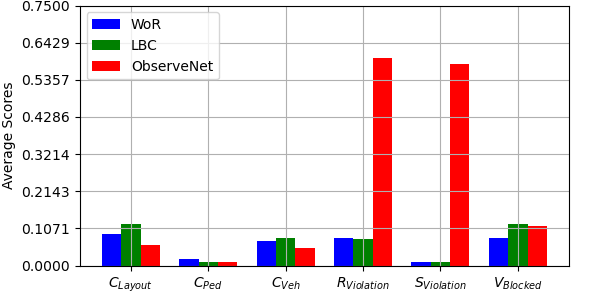}
        \vspace{-2em}
		\caption{\textbf{Experimental results on CARLA Leaderboard~\cite{A}.} Average number of infractions for all tested routes. From left to right: layout collisions ($C_{Layout}$), pedestrian collisions ($C_{Ped}$), vehicle collisions ($C_{Veh}$), red light violations ($R_{Violation}$), stop sign violations ($S_{Violation}$) and blocked vehicle violations ($V_{Blocked}$).}
        \label{fig:collision_statistics}
	\end{center}
    \vspace{-1em}
\end{figure}

Since our focus is on eliminating collisions between the ego-vehicle and other traffic participants, we aim to optimize the number of collisions, route deviations and route completion. We have conducted a total of $40$ routes, each repeated three times for each evaluated algorithm, comparing side by side collision metrics and other infractions, as shown in Fig.~\ref{fig:collision_statistics}. Although the number of infractions like stop sign violations and red light violations are more frequent in our approach, the average number of collisions remained overall lower than for LBC and WOR.

From Fig.~\ref{fig:collision_statistics} it can be observed that LBC and WOR tend to perform better than our approach for stop sign and red light violations. This is happening because we use range sensing only, where stop signs and traffic lights are not recognized.

\subsection{Ablation Study}

\begin{figure}
\centering
	\begin{center}
		\includegraphics[scale=0.92]{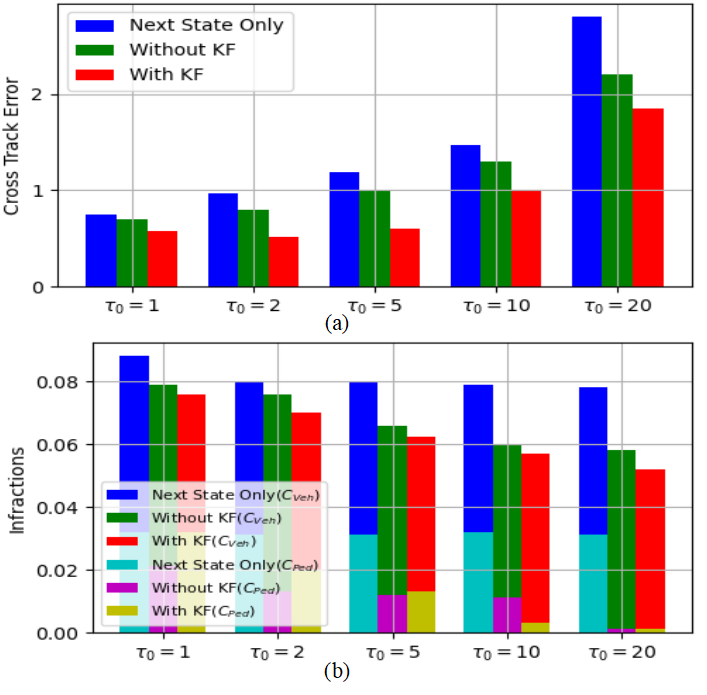}
        \vspace{-1em}
		\caption{\textbf{Ablation study results.} The Kalman filter was switched on/off, while the prediction horizon was set to $1s$, $2s$, $5s$, $10s$ and $20s$, respectively.}
        \label{fig:ablation_study}
	\end{center}
    \vspace{-2em}
\end{figure}

In order to asses the core components of the algorithm, we have performed the following ablations to the ObserveNet structure from Fig.~\ref{fig:network_architecture}: \textit{i}) ablation of the Kalman filter, \textit{ii}) ablation of the prediction horizon and \textit{iii}) control based on the prediction of the next vehicle's state. In the last case, the Kalman filter and the temporal planner were removed, together with the $[t+2, t+\tau_o]$ branches. The kept $t+1$ branch was trained to predict only the next state of the vehicle based on $[t-\tau_i, t]$ observations. Fig.~\ref{fig:ablation_study}(a) shows abblations against the cross-track error $e_{ct}$, while Fig.~\ref{fig:ablation_study}(b) takes into consideration pedestrian and vehicle collisions.

The cross-track error $e_{ct}$ is proportional to the length of prediction horizon $\tau_o$. The further in time we predict observations, the larger the cross-track error is. Our intuition for this increase in $e_{ct}$ is that ObserveNet is anticipating the visual dynamics of the obstacles encoded in the raw measurements, thus optimizing the vehicle's trajectory against collisions, while diverging from the reference path. This phenomenon can be seen in Fig.~\ref{fig:ablation_study}(b), where the pedestians and other vehicles collisions are decreasing proportional to the length of $\tau_o$.

Without the smoothing property of the Kalman filter, the values of the control signals are higher, leading to a larger error. A decrease in performance is observed when the vehicle is controlled solely by predicting its next state.

\subsection{RovisLab AMTU Experiments}

In addition to the CARLA simulated experiments, we have tested ObserveNet and DWA on real-world navigation tasks using the RovisLab AMTU robot from Fig.~\ref{fig:model_car}. Due to its better performance over LBC and WOR, we have chosen to conduct real-world experiments only with ObserveNet as the learning-based controller and DWA. RovisLab AMTU is an AgileX Scout 2.0 platform which acts as a 1:4 scaled car, equipped with a $360^{\circ}$ Hesai Pandar 40 Lidar, 4x e-130A cameras providing a $360^{\circ}$ visual perception of the surroundings, a VESC inertial measurement unit, GPS and an NVIDIA AGX Xavier board for data processing and control. The robot navigated indoor and outdoor environments, while avoiding dynamic obstacles.

The prediction horizon $\tau_o$ was set to $5s$ based on trial-and-error experiments, while $10km$ of training data was acquired for the indoor and outdoor experiments each. Both real-world experiments have been conducted at Transilvania University's research institute, amounting to $38.000$ pairs of Lidar and vehicle state data.

We have replicated similar scenarios as the ones performed in simulation, with the difference that the moving vehicles from CARLA have been replaced by persons interacting with the robot in similar ways as in the five testing schemes described above. Table~\ref{tab:results_rovislab_amtu} shows quantitative results on the real-world experiments. The results show an increased improvement in tracking accuracy and obstacle avoidance in the case of ObserveNet Control.

\begin{table}[]
    \centering
    \resizebox{\columnwidth}{!}
    {
    \begin{tabular}{ccccccc}
        \hline
        \textbf{Scenario} & \textbf{Algorithm} & \textbf{$N_C$} & \textbf{$\bar{e_{ct}}$} & \textbf{$\bar{e_{\phi}}$} & $\bar{e_{a}} / \bar{e_{\delta}}$ & \textbf{$C_{100\%}$}\\
        \hline
        \multirow{2}{*}{Overtake}           & DWA  & 1 & \textbf{0.27} & 0.055 & 0.0022 / 0.0022 & 100 \\
                                            & Ours & \textbf{0} & 0.55 & \textbf{0.001} & \textbf{0.0004} / \textbf{0.0004} & 100 \\
        \hline
        \multirow{2}{*}{Opposite}           & DWA  & 1 & 1.32 & 0.006 & 0.005 / 0.002 & 100 \\
                                            & Ours & \textbf{0} & \textbf{0.19} & \textbf{0.002} & \textbf{0.0007} / \textbf{0.0007} & 100 \\
        \hline
        \multirow{2}{*}{Side cut-off}       & DWA  & 1 & 0.12 & 0.1 & 0.003 / 0.004 & 100 \\
                                            & Ours & \textbf{0} & \textbf{0.09} & \textbf{0.007} & \textbf{0.001} / \textbf{0.001} & 100 \\
        \hline
        \multirow{2}{*}{Rear pass-by}       & DWA  & N/A & 2.23 & \textbf{0.06} & 0.009 / \textbf{0.0025} & 100 \\
                                            & Ours & N/A & \textbf{0.72} & 0.09 & \textbf{0.006} / 0.006 & 100 \\
        \hline
        \multirow{2}{*}{Multi-obstacles}      & DWA  & 3 & \textbf{0.035} & 0.08 & 0.009 / 0.008 & 70.32 \\
                                            & Ours & \textbf{0} & 0.067 & \textbf{0.01} & \textbf{0.005} / \textbf{0.005} & 100  
    \end{tabular}
    }
    \caption{Quantitative results on RovisLab AMTU.}
    \label{tab:results_rovislab_amtu}
    \vspace{-3em}
\end{table}



\section{Conclusions}
\label{sec:conclusions}

This paper introduces ObserveNet Control, which is an observation prediction control method for autonomous vehicles composed of a DNN for observations predictions and a temporal path planner. We have shown that our control approach has an improved path tracking and obstacles avoidance accuracy when augmented with predicted observations. Additionally, the training of the system is self-supervised from driving recordings, which requiring manual human annotations. Because we only predict raw observations, one drawback of ObserveNet is that it lacks driving context, hence resulting in a large number of traffic violations. This can be easily corrected by integrating our method into autonomus driving systems that use visual data to detect additional cues, such as lanes,  or traffic signs. In such a case, ObserveNet would contribute with its raw observations predictions to the reconstructed virtual environment model used by the vehicle to plan its path. As future work, we plan to increase the time prediction horizon, as well as its possible application in mobile robotics and simultaneous localization and mapping.


\bibliographystyle{IEEEtran}
\bibliography{references}

\end{document}